\documentclass{article}

\PassOptionsToPackage{numbers, compress}{natbib}
\usepackage[preprint]{nips_2018}

\usepackage[utf8]{inputenc} % allow utf-8 input
\usepackage[T1]{fontenc}    % use 8-bit T1 fonts
\usepackage{hyperref}       % hyperlinks
\usepackage{url}            % simple URL typesetting
\usepackage{booktabs}       % professional-quality tables
\usepackage{amsfonts}       % blackboard math symbols
\usepackage{nicefrac}       % compact symbols for 1/2, etc.
\usepackage{microtype}      % microtypography
\usepackage{graphicx}
\usepackage{subfigure}
\usepackage{makecell}
\usepackage{multicol}
\usepackage{caption}
\newcommand{\PropSystem}{d-DeVIS}

\title{A Gray Box Interpretable Visual Debugging Approach for Deep Sequence Learning Model}

\author{
    \And
    Md Mofijul Islam$^1$, Amar Debnath$^1$, Tahsin Al Sayeed$^1$, Jyotirmay Nag Setu$^1$,\\
    \textbf{Md Mahmudur Rahman$^1$,Md Sadman Sakib$^1$, Md Abdur Razzaque$^1$, Md. Mosaddek Khan$^1$}\\\textbf{Swakkhar Shatabda$^2$}\\
    \{akash.cse.du,amar.csedu,tahsinalsayeed,jyotirmaynag.cse,mrr.rana13,sakib.csedu21\}@gmail.com\\
    razzaque@du.ac.bd,mosaddek@du.ac.bd,swakkhar@cse.uiu.ac.bd\\
    $^1$Department of CSE, University of Dhaka\\
    $^2$Department of CSE, United International University
}
\begin{document}

\maketitle

\begin{abstract}
 Deep Learning algorithms are often used as black box type learning and they are too complex to understand. The widespread usability of Deep Learning algorithms to solve various machine learning problems demands deep and transparent understanding of the internal representation as well as decision making.
  Moreover, the learning models, trained on sequential data, such as audio and video data, have intricate internal reasoning process due to their complex distribution of features. Thus, a visual simulator might be helpful to trace the internal decision making mechanisms in response to adversarial input data, and it would help to debug and design appropriate deep learning models. However, interpreting the internal reasoning of deep learning model is not well studied in the literature. In this work, we have developed a visual interactive web application, namely \PropSystem, which helps to visualize the internal reasoning of the learning model which is trained on the audio data. The proposed system allows to perceive the behavior as well as to debug the model by interactively generating adversarial audio data point. The web application of \PropSystem $\:$is available at \url{ddevis.herokuapp.com}.
\end{abstract}

\section{Introduction} \label{sec:Introduction}

Machine Learning(ML) algorithms have been pouring the blessings in a form of solving Artificial Intelligence(AI) problems, such as classification, clustering, genomics data visualization, etc. Deep Learning(DL), an influential extension of ML, has been evolving rapidly in recent years and successfully being applied in solving various real-world problems including machine translation, speech recognition, image classification, etc\cite{machinetrans}\cite{speechrecog}\cite{imagenet}. While traditional ML models require external domain knowledge, DL is mostly characterized for efficient learning of the non-linear complex feature representation without having domain expertise. Hence, the DL model remains as a black-box type learning for practitioners and researchers. In effect, the interpretability and transparency of DL models have been reduced significantly \cite{visualSurvey}. Although DL approaches have been studied widely, a few works address the interpretability issue of deep learning models in the literature.

With the increasing use of the DL methodologies in real-world systems, such as self driving car and medical imaging, it becomes a prime concern to have publicly understandable systems explaining the underlying reasoning. Although the linear systems can be easily demonstrated with simple examples having mathematical proofs, non-linear systems, such as Deep Neural Network(DNN), is complex to understand and visualize. Nonetheless, the general users as well as researchers need to understand the mechanism of the algorithms to debug and determine appropriate learning model. In addition, the teachers and the learners are interested to visualize the algorithms to develop the basic intuition of the algorithm. The researchers have been working to utilize the visualization approaches to teach the ML algorithms \cite{nvis} while it has been proven that people can grasp the principles of an algorithm better when they are taught using visualization approaches \cite{LearnTeachProg}\cite{InternalStateProfTough}.

\par Visualization of internal operation details of a machine learning algorithm has been studied previously in \cite{1992visualizing}, where the authors have surveyed several visualization techniques to understand the learning and decision-making processes of neural networks and also describe their work in knowledge-based neural networks. After the explosion of deep learning applications in computer vision and machine translations, researchers have been trying to visualize the interpretations of the specialized algorithms used for different kinds of unstructured data. In \cite{cnn_vis_pre}, authors have introduced a novel visualization technique
that gives insight into the function of intermediate feature layers and the operations of Convolutional Neural Network(CNN). Nonetheless, it is rather black-box type visualization approach to reveal the model behavior, as such it can not interpret the internal reasoning. In \cite{shapeshop}, authors have developed an interactive system to enable users understand and explore the deep learning models and get an insight on the learning mechanisms of image classifiers. It introduces a gray-box type approach but does not demonstrate how classifiers work in response to sequence audio data. 

\par In this paper, we have designed a deep Sequence Learning Model Debugger and Visual Interactive Simulator, namely \PropSystem, that focuses on gray box concept, where outcome of an internal block is transparent to the users. More explicitly, we are interested to visualize the internal feature representation of a deep sequence learning model (i.e. CNN) in response to multi modal audio sequence data. The layer wise visualization of hidden features in \PropSystem $\:$ assists us to understand the interpretation of feature extraction methods of DL models. The main contributions of the paper are as follows:

\begin{itemize}
\item A web-based application, \PropSystem, to visualize the representation of hidden layers' features and the behavior of the CNN model in response to the adversarial audio sequence data.
\item \PropSystem, allows user to interactively change the audio features, such as pitch, amplitude etc, and interpret the behavior of the learning model based on the modified data.
\item We have designed a visually transparent debugging User Interface(UI), which demonstrates layer-wise features' representation and model hyper parameters. In so doing, it guides DL model's debugging.

%\item \PropSystem enables users to change the hyper-parameters of the DL model, such as kernel size, hidden layers weights and filter size. After that, user can re-trained the updated model and examine the updated model behavior. Except this model re-training feature we deployed other features on web app, as it incurs a lot of computation resources. However, we have the plan to open sourced the complete project the code. 

\item \PropSystem $\:$ enables users to hear and visualize the intermediary hidden layer results, layer-wise converted audio outputs and weight distributions, in order to interpret the final prediction. It also allows practitioners to compare the performance of the learning model in response to different adversarial audio input.   
\end{itemize}

The rest of the paper is structured as follows. In Section \ref{sec:RelatedWork}, we discuss the related work. Thereafter, Section \ref{sec:ProposedSystem} is focused on the goals and features of the proposed system. Section \ref{sec:UseCases} describes the use cases of \PropSystem. Finally, Section \ref{sec:Conclusion} concludes with future plans.
\section{Related Work} \label{sec:RelatedWork}

The recent widespread use of deep learning models in various artificial intelligence task attracts both the visualization and the deep learning communities to deal with the new challenge of improving the interpretability and explainability of these models \cite{1992visualizing}. It is worth mentioning that visualizing the Neural Network (NN) models is not a new research domain. To be precise, it has been studied well before the recent surge of deep learning models. For instance, N$2$Vis \cite{nvis} visualizes the attributes of NN, such as hidden layers weights, weights' volatility, network structure and nodal activation levels. Nonetheless, most of the previous approaches utilize the static graphical visualization to describe hidden reasoning of the learning models.

In recent years, a number of works have been sought to address the explainability and transparency issue of the DL models and few others have been  focused on designing interactive visualization models to illustrate underline reasoning. For example, Tensorflow Playground \cite{tensorPlayground} designed an interactive interface, where users can change the parameters and structure of the NN models and examine their effect. Moreover, ShapeShop \cite{shapeshop} enables the users to interactively change input image and visualize the behavior and feature's representation of the DL models. Similarly, in \cite{imageClassification}, authors designed an application, which allows an user to examine the behavior of a DL based image classifier.

\par Apart from these black-box visualization approaches, a number of works visualize the behavior of deep learning models. For instance, in \cite{karpathy2015visualizing}, authors present a static visualization of hidden state representation and the prediction model behavior of Long-Short-Term-Memory(LSTM) based language model. Similar to the previous work, LSTMVis \cite{lstmvis} designed an interactive visualization approach to visualize the hidden state representations of recurrent neural network and allows user to examine the internal behavior of LSTM model on different application scenarios. Additionally, in \cite{cnn_vis} and \cite{cnn_vis_pre}, authors visualize the Convolutional Neural Network (CNN) and provide visually explainable reasoning of internal feature representation. Furthermore, Seq2Seq \cite{seq2seq} designed a visual debugging tools for the sequence-to-sequence learning model and enables users to interact with the model to develop an insight about the model.

\par Inspired from the previous works done by \cite{shapeshop,lstmvis,seq2seq}, we have designed an interactive visual DL models debugging system, \PropSystem: Deep Sequence Learning Model's Debugger and Visually Interactive Simulator. Most of the previous works utilize the black box visualization approaches to help developing the basic intuition of the deep learning models. Surprisingly, visualizing the deep learning model behavior and features representation of the multimodal data, such as audio or video, is neglected in the literature. Moreover, visualizing the correlation between the hidden layer features representation and the model behavior is not properly studied for sequence models. \PropSystem $\:$ allows user to interactively change the multimodal audio data to generate adversarial data examples and enables users to examine the deep learning model behavior to visualize the features representation. 
\section{Design and Development of \PropSystem} \label{sec:ProposedSystem}
In this section, we present the key components and goals for designing our proposed interactive application to visualize DL model in response to the adversarial data input. 
We take into considerations the interactivity of the users and flexibility of the system. To do so, we have developed a web application that shows the gray box debugging method for deep neural network of sequence data. The prime goal of designing \PropSystem $\:$ is to make the learning and debugging DL model user friendly and also ensure that it should be able to visualize the internal reasoning of deep sequence model and features representation of hidden layers with the help of an interactive user interface. Table~\ref{tab:goals} lists a number of major design goals for designing an interpretable deep audio sequence learning model.

\begin{table}[!h]
	\caption{Design Goals of \PropSystem}
	\label{tab:goals}
	\centering
	\begin{tabular}{ p{3.5cm} |p{9.5cm}}
		\toprule
		\multicolumn{1}{c|}{Goals} & 
		\multicolumn{1}{c}{Description}     \\
		\midrule
		G1: Improve DL Models Interpretability and Transparency & An interpretable system of DL models depicts how deep sequence learning models work and how the hidden layer features can help to easily interpret the functionality of the learning model.     \\
		\hline
		G2: Gray-Box Visual Debugging & A good grasp of the feature extraction method of deep neural networks is required for DL enthusiast and \PropSystem $\:$ provides a
		fluid gray box debugging experience which enables the users to understand how the features of the hidden layers affect the training.    \\
		\hline
		G3: Interactively Examining the Deep Sequence Model Behavior & An interactive tool is required, where user can manipulate audio features(such as slicing, cross-fading, repetition, etc) to generate adversarial example data. Moreover, it allows user to examine the internal reasoning in response to the modified adversarial data.\\
		\hline
		G4: Comparison and exposure of the extracted features from audio data & The proposed system must enable users to listen the extracted audio data from different layer after applying CNN filters. Hence, users should be able to grasp the extracted hidden layer audio features. \\
		\bottomrule
	\end{tabular}
\end{table}

\subsection{Features of \PropSystem}
We have designed \PropSystem $\:$ as an interactive web application while considering the design goals listed in Table~\ref{tab:goals}. The primary goal of our proposed system is to ease the interpretation of the intermediate reasoning and the deep audio sequence learning model. We divide the proposed \PropSystem $\:$ model into the following three major components.

\begin{itemize}
	\item Model Visualization.
	\item Audio Feature Manipulation
	\item Adversarial Feature Comparison
\end{itemize}

\subsubsection{Model Visualization}

\begin{figure*}[!ht]
	\includegraphics[width=\textwidth,height=5.5cm]{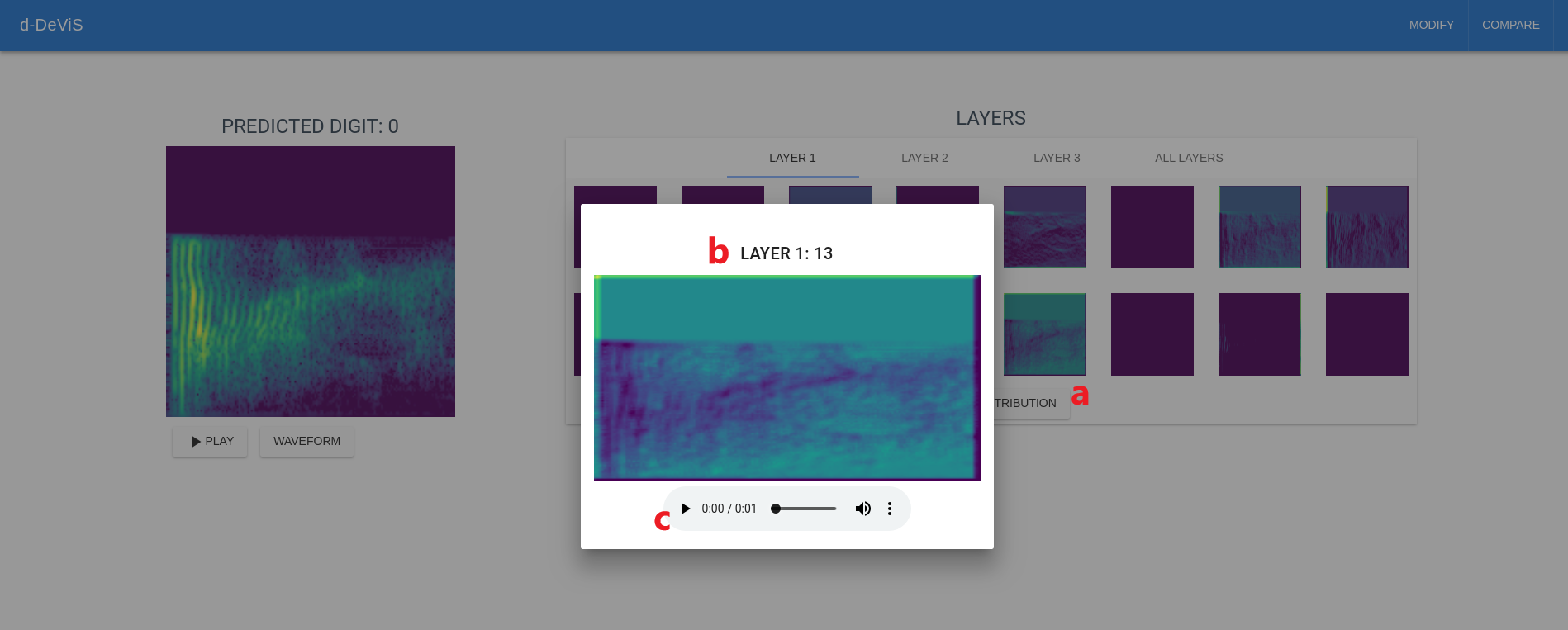}
	\caption{Visualization of audio features extracted by CNN layer filter.}
	\label{fig:layer}
\end{figure*}

The primary purpose of our work is to interpret the internal reasoning of the deep sequence learning model in response to adversarial audio example data. For this reason, \PropSystem  $\:$ provides an interactive web application interface, which depicts the intermediate layer wise visual features representation in the form of audio spectrogram. Moreover, we employed the inverse Fourier transformation to extract the audio features from the intermediate layer spectrogram. \PropSystem $\:$ allows user to not only visualize the features extracted by the hidden layer filters, but also it enables them to listen to the audio representation of the features of the input audio extracted by the CNN. The web interface to visualize the layer wise feature is depicted in Fig~\ref{fig:layer}. Furthermore, \PropSystem $\:$ allows the users to examine the weight distributions of the internal hidden layer. To extract the intermediate features, we trained a baseline CNN model on audio sequence data. The details of the trained model and the backed system of \PropSystem $\:$ is presented in Section~\ref{sec:implementation_of_ddevis}.

During any forward propagation step, the spectrogram feature data of the audio files are traversed through the hidden layer of the CNN. At each layer, the convolution filter tries to extract significant hidden features from the audio data input and optimizes itself during backward propagation in order to minimize the training loss. In our system, the users will be able to upload an audio file or record an audio of their own. After the processing of the input, our system will calculate the logarithmic spectrogram and feed it into the trained model to produce the prediction. At each convolution layer there are predefined tunned filters. The types of features CNN] extracts from the input data depends on the filters. In our trained model, the first layer and second layers have 16 filters, the third layer has 32 filters, each. So, our system visualizes the features corresponding to the filters and also the distributions of the trained weights.

\par A particular feature extracted by the 13th filter of first layer is visualized in Fig ~\ref{fig:layer}(a). When a user clicks on the image it zooms in to show the spectrogram clearly. Users can also listen to the hidden extracted feature by clicking on the play button, which is depicted in Fig.~\ref{fig:layer}(c).

\subsubsection{Adversarial Feature Comparison}

\begin{figure*}[ht]
	\includegraphics[width=\textwidth,height=7cm]{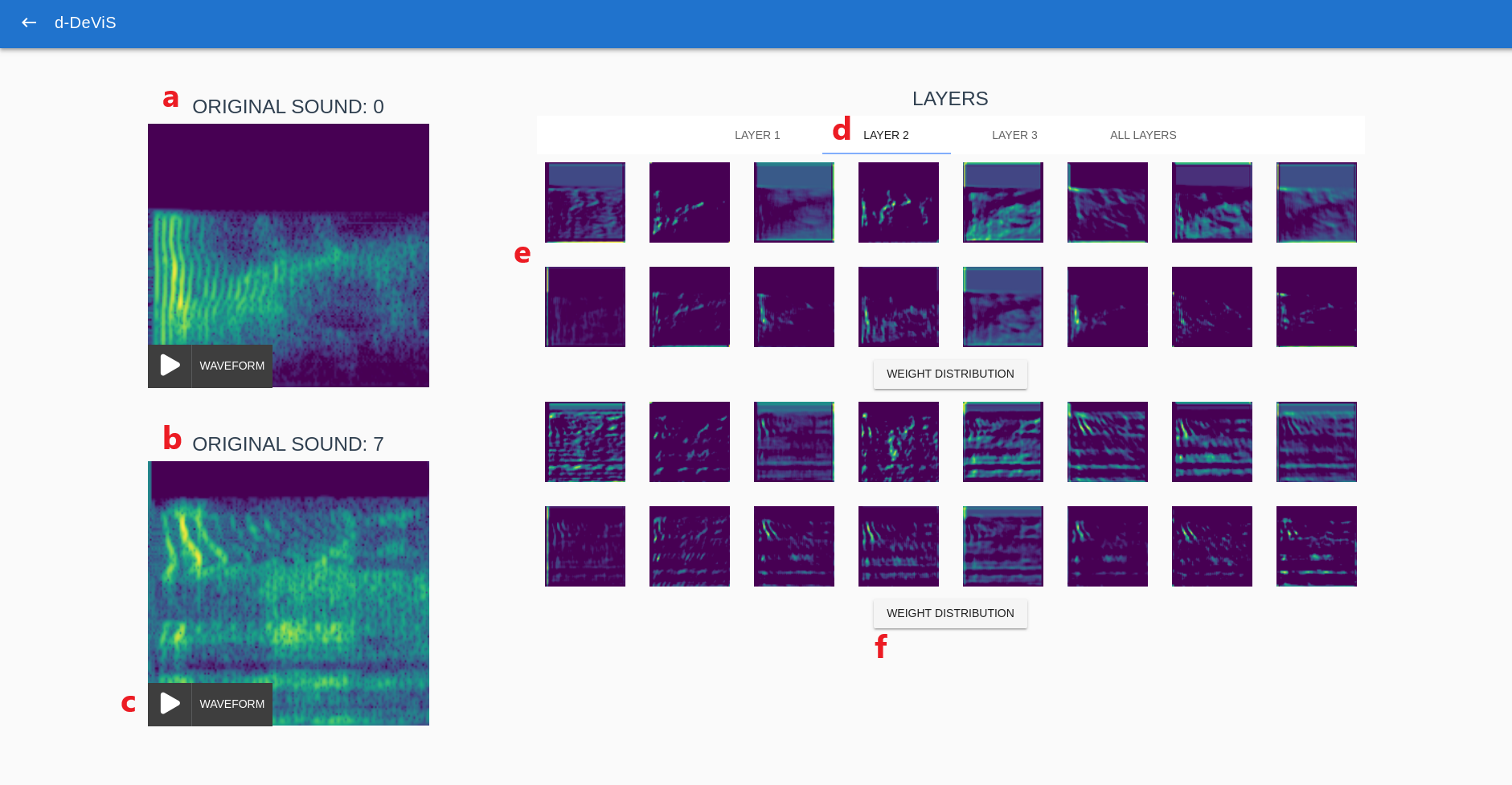}
	\caption{Comparison of different audio inputs.}
	\label{fig:comp}
\end{figure*}

\PropSystem $\:$ allows users to interpret the DL model behavior by examining the intermediate feature representation based on the different audio data input. The adversarial behavior comparison is illustrated in Fig~\ref{fig:comp} and the different module of this feature is presented in red alphabets. The module a and b are the two spectrogram representation of the two audio inputs with their predictions by the trained deep sequence learning model. Users can observe the feature representation of different layers in module d and e. There are different spectrogram images of the extracted features and users can click on them to listen to the audio representation. Finally users can also see the weight distribution of each layers by clicking on the button marked by f.  

\subsubsection{Audio Feature Manipulation}
Our proposed interactive system, \PropSystem, enables the users to not only examine the behavior of the learning model in response to a sample file or recording of their voice but also it allows users to manipulate the different properties of audio example data and thus enables to generate adversarial example data. Among the various characteristics of the audio, which can be changed, \PropSystem $\:$ allows to manipulate the following audio features. 

\begin{itemize}
	\item \textbf{Slicing} allows users to slice an audio.
	
	\item \textbf{Cross-fading} changes the amplitude of the sound waves.
	
	\item \textbf{Changing the loudness} option will make the beginning louder and the ending quieter.
	
	\item \textbf{Repeating} option repeats the sound twice.
	
	\item \textbf{Invert:} allows to invert the sound wave, i.e. inverted sound will be played from the ending.
	
	\item \textbf{Fade:} option fades in for a particular time and then fades out similarly.
	
\end{itemize}

\begin{figure*}[!h]
    \subfigure[Original Sound]{\includegraphics[width=\columnwidth]{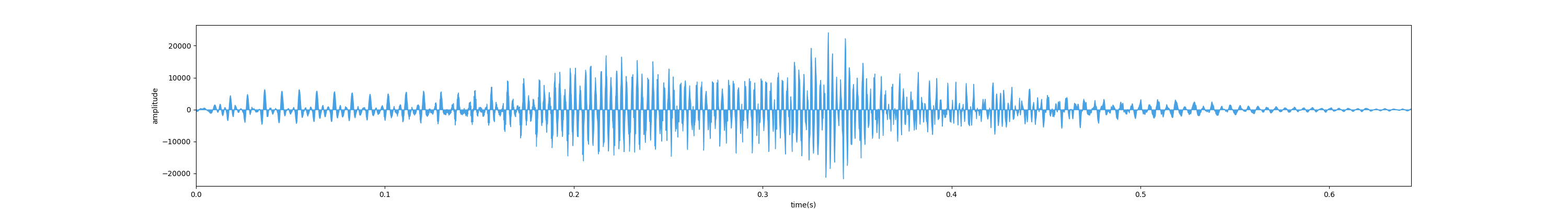}}
	\subfigure[Slicing audio]{\includegraphics[width=0.5\columnwidth]{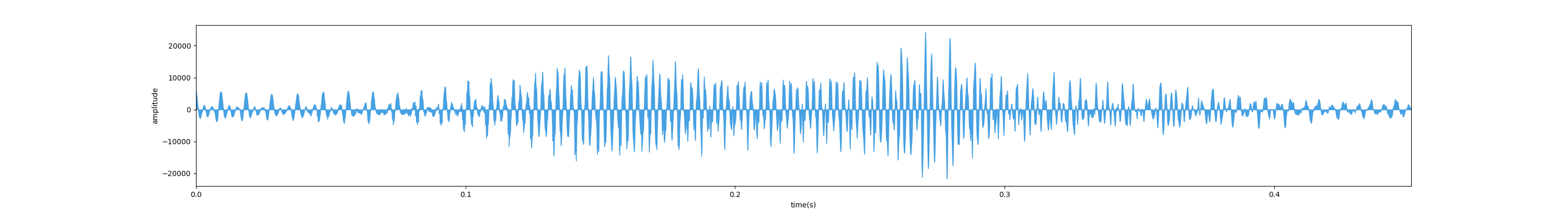}}
	\subfigure[Cross-fading]{\includegraphics[width=0.5\columnwidth]{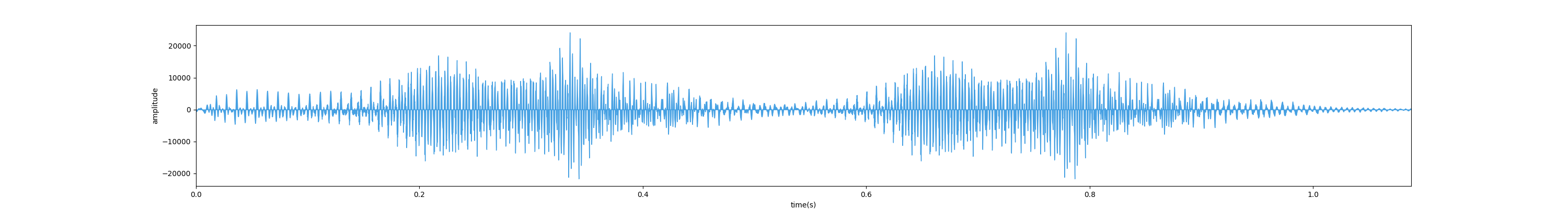}}
	\subfigure[Changing loudness]{\includegraphics[width=0.5\columnwidth]{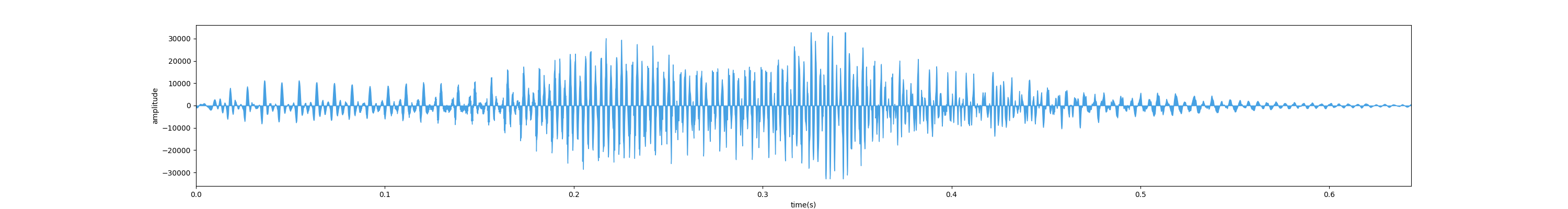}}
	\subfigure[Repeating sound]{\includegraphics[width=0.5\columnwidth]{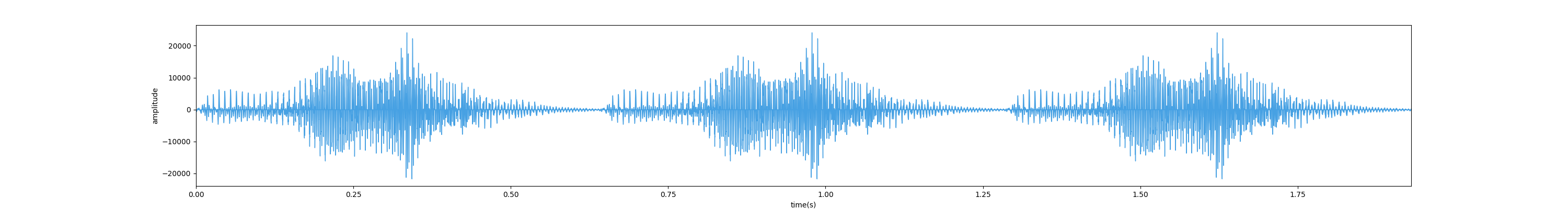}} 
	\subfigure[Invert]{\includegraphics[width=0.5\columnwidth]{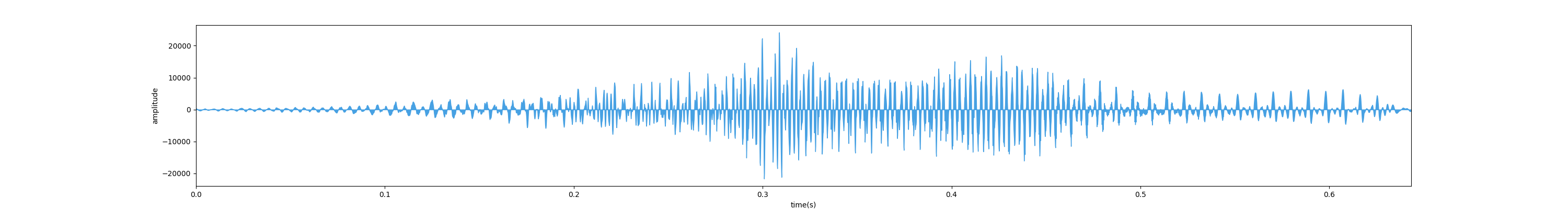}} 
	\subfigure[Fade]{\includegraphics[width=0.5\columnwidth]{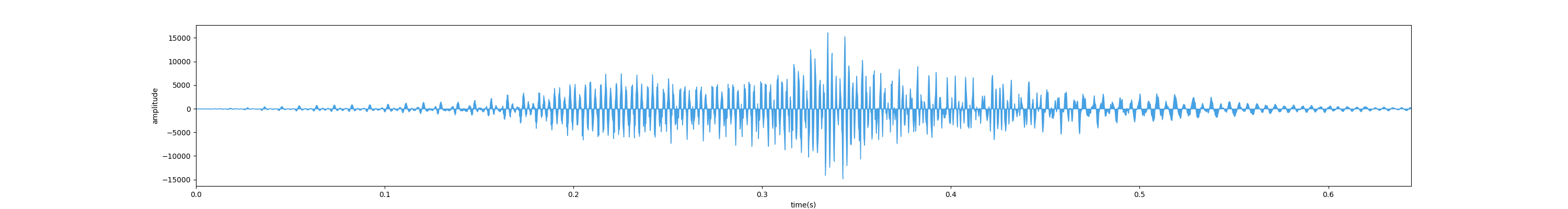}} 
	\caption{Visualize the modified audio features. (\textit{time} vs \textit{amplitude})}
	\label{fig:editFeatures}
\end{figure*}

A pictorial modification of audio feature is presented in Fig.~\ref{fig:editFeatures}. After manipulating and generating the audio example data, \PropSystem $\:$ allows users to examine the behavior of audio deep sequence learning model by observing behavior changes in response to the original and adversarial audio data input.

\par In Table \ref{tab:goalsMapping}, we discuss all the features of \PropSystem $\:$ and how the features meet the design goals.
\begin{table}[!h]
	\caption{Mapping of \PropSystem $\:$ features and goals }
	\label{tab:goalsMapping}
	\centering
	\begin{tabular}{ p{12cm} |p{1.5cm}}
		\toprule
		\multicolumn{1}{c |}{Feature} & 
		\multicolumn{1}{c}{Goal}\\
		\midrule
		\textbf{Visualizing the Hidden Extracted Features of Convolutional neural networks:} \PropSystem $\:$ provides visualization of the extracted features in each layer as image data and shows the various features of different filters of the deep Convolutional Neural Network. Users can also hear the audio representation of the hidden extracted features & G1 \& G2 \& G4\\
		\hline
		
		\textbf{Interactive User Experience:} For a fluid user experience,we provide an interactive platform for the users so that they will be able to focus on the productivity of the system without any unnecessary hassle. & G3 \& G4\\
		\hline
		\textbf{Visualizing the Audio Features as well as Modifying the Waveforms:} Due to the complex structure of audio data, our system let's users modify various aspects of the sound property and visualize the updated waveform to provide a keen knowledge on audio data representation. & G2 \& G3\\
		\hline
		\textbf{Custom Audio Input for Testing and Feature Distribution Visualization:} User can not only upload a default audio data but also they can record custom speech to test the trained model. Proper distribution of the weights is also visualized.  & G1 \& G3 \\
		\hline
		\textbf{Comparing different audio inputs and their hidden features:} \PropSystem $\:$ also enables users to measure the differences of different audio inputs and check their extracted layer features. & G4  \\
		\bottomrule
	\end{tabular}
\end{table}

\subsection{Implementation of \PropSystem}
\label{sec:implementation_of_ddevis}
\PropSystem $\:$ is developed as a web application so that users can seamlessly interact with the system to interpret the behavior of deep learning model by generating adversarial audio input data.  In the following section we present the implementation details of \PropSystem. The source code of our implementation is available at \url{https://github.com/anon-conf/d-DeVIS}.

\subsubsection{Trained Deep Learning Model with Audio Data}
\begin{figure*}[!ht]
	\centering
	\includegraphics[width=0.8\textwidth]{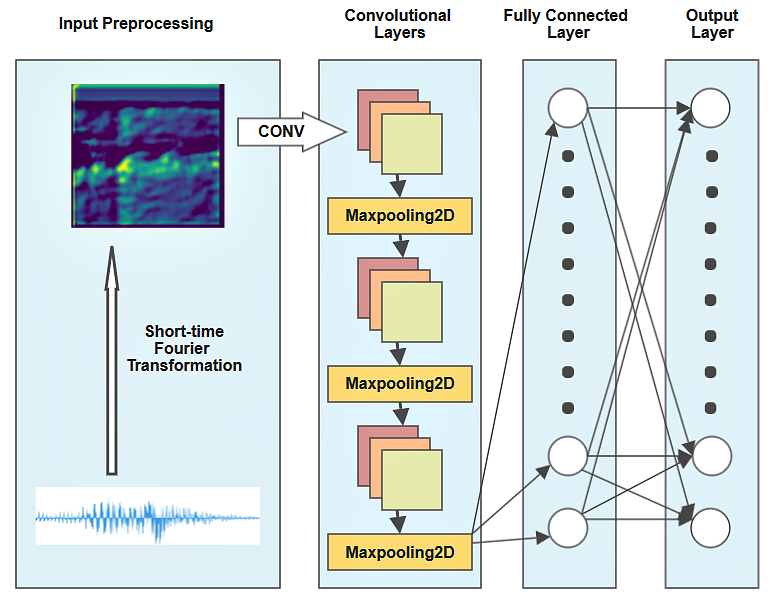}
	\caption{Convolutional Neural Network Architecture of \PropSystem.}
	\label{fig:algo}
\end{figure*}

We trained a CNN model on Speech Commands dataset \footnote{\url{www.kaggle.com/c/tensorflow-speech-recognition-challenge}}, which is used to visualize the behavior of the model. The dataset consists of almost 30 speech classes but for the sake of the simplicity and reduction of training time we used 10 classes, which are the audio recordings of zero to nine digits in English language. All the clips are one second long. We calculated logarithmic spectrogram as features of the audio (.wav) data to feed into the training model. A three layer Convolutional Neural Network (CNN) is used as the spectrogram feature matrix represents an image and CNNs have proven to be decent at image classification. The Convolutional Architecture consisted of 3s set of filters with different square kernel of sizes [7,5,3]. We used max pooling after every filter to reduce the sizes of the output matrices and added necessary dropout to reduce overfitting. The complete architecture of the training model is depicted in Fig ~\ref{fig:algo}. Our baseline model reached 95\% validation accuracy with a minimal hyper parameter tuning.
%CNN model parameters
We have used Keras deep learning framework which is a wrapper library of Tensorflow to train our deep learning model. We have utilized the computation system of Google Colaboratory platform for the training purpose.

%\subsubsection{Extract Hidden Layer Features}  

\subsubsection{Front-end and Back-end of d-DeVIS}

Audio files manipulations such as Slicing the audio, Changing Loudness, Cross-fading, Repeating the Sound, Invert and Fade are done using Numpy, Pydub and Scipy libraries. Matplotlib is utilized to visualize the audio features. After the training of the model, we have saved the data using Pickle python module. We have used HTML5, CSS, Javascript for designing the front-end of the application. We used the Vue.js framework to build an SPA and communicated with the server using REST API. The back-end is built in python using the Flask framework. 
\section{Use Cases} \label{sec:UseCases}

While experimenting with the system, we have applied several modifications to both the CNN trained model and the input audio file. Then, we have analyzed the obtained results by tuning with the system. In the remainder of this section, we present important use cases that demonstrate the general applicability of our system. A demo video of our proposed system \PropSystem $\:$ can be found at \url{http://bit.ly/ddevis-demo}.

\begin{itemize}
	\item \textbf{Visualizing the Audio Features:}
	Speech is a sequence data which is hard to grasp just by looking at the amplitude vs time representation. In our system, a user can upload or record a customized audio file and tune with various aspects of the waveform. Therefore, predictions of the audio will change with accordance to the change in the waveforms and users can easily observe the changed results.
	
	\item \textbf{Learning Medium for the Academia:}
	Our system provides an interactive web application with which learners will be able to test various types of aspects of audio data and the deep learning model. By using \PropSystem $\:$, academics can provide appropriate insights of the feature extraction method of neural networks to the students. Hence, it can be a great medium for learning.
	
	\item \textbf{Experimenting Platform for AI Enthusiasts:}
	We provide a platform for easy training and proper results of the feature extractions which are shown as a form of images. Users can test their own custom input and observe the decisive hidden features that make the distinctions between the inputs. Thus, the feature manipulation and interactivity of the system will inspire the deep learning enthusiasts and engineers to do various experiments on it.
\end{itemize}

\section{Conclusion} \label{sec:Conclusion}

%{\tiny In its current form, this is more like a summary than a conclusion.They are not the same. Please add a couple of sentences at the beginning of the conclusion, which gives a reader clear idea about your contributions with respect to the state-of-the-art. The purpose is to give a reader an opportunity to skim the paper by reading Abstarct and Conclusion only.}

\PropSystem $\:$ allowed the users to visualize how CNN recognizes digits from audio sequence data. It collected input from user and allowed them to interactively manipulate it. The tool easily allowed the comparison of the given input with other adversarial examples. Overall, this helped users to develop a better intuition of the underlying reasoning of the model which allowed them to make more learned decisions regarding learning model development.

\par In future extension of \PropSystem, we have the plan to visualize other sequence deep learning models behavior and allow users to manipulate the input data representation interactively. Moreover, visualizing the hidden layer complex feature representations for multi-modal sequence data is a great avenue for future research work.

\bibliographystyle{unsrt}
\bibliography{bibs}

\begin{thebibliography}{10}

\bibitem{machinetrans}
Y.~Wu, M.~Schuster, Z.~Chen, Q.~V. Le, M.~Norouzi, W.~Macherey, M.~Krikun,
  Y.~Cao, Q.~Gao, K.~Macherey, {\em et~al.}, ``Google's neural machine
  translation system: Bridging the gap between human and machine translation,''
  {\em arXiv preprint arXiv:1609.08144}, 2016.

\bibitem{speechrecog}
D.~Amodei, S.~Ananthanarayanan, R.~Anubhai, J.~Bai, E.~Battenberg, C.~Case,
  J.~Casper, B.~Catanzaro, Q.~Cheng, G.~Chen, {\em et~al.}, ``Deep speech 2:
  End-to-end speech recognition in english and mandarin,'' in {\em
  International Conference on Machine Learning}, pp.~173--182, 2016.

\bibitem{imagenet}
A.~Krizhevsky, I.~Sutskever, and G.~E. Hinton, ``Imagenet classification with
  deep convolutional neural networks,'' in {\em Advances in neural information
  processing systems}, pp.~1097--1105, 2012.

\bibitem{visualSurvey}
F.~M. Hohman, M.~Kahng, R.~Pienta, and D.~H. Chau, ``Visual analytics in deep
  learning: An interrogative survey for the next frontiers,'' {\em IEEE
  Transactions on Visualization and Computer Graphics}, 2018.

\bibitem{nvis}
M.~J. Streeter, M.~O. Ward, and S.~A. Alvarez, ``Nvis: An interactive
  visualization tool for neural networks,'' in {\em Visual Data Exploration and
  Analysis VIII}, vol.~4302, pp.~234--242, International Society for Optics and
  Photonics, 2001.

\bibitem{LearnTeachProg}
A.~Robins, J.~Rountree, and N.~Rountree, ``Learning and teaching programming: A
  review and discussion,'' {\em Computer science education}, vol.~13, no.~2,
  pp.~137--172, 2003.

\bibitem{InternalStateProfTough}
B.~Du~Boulay, ``Some difficulties of learning to program,'' {\em Journal of
  Educational Computing Research}, vol.~2, no.~1, pp.~57--73, 1986.

\bibitem{1992visualizing}
M.~W. Craven and J.~W. Shavlik, ``Visualizing learning and computation in
  artificial neural networks,'' {\em International journal on artificial
  intelligence tools}, vol.~1, no.~03, pp.~399--425, 1992.

\bibitem{cnn_vis_pre}
M.~D. Zeiler and R.~Fergus, ``Visualizing and understanding convolutional
  networks,'' in {\em European conference on computer vision}, pp.~818--833,
  Springer, 2014.

\bibitem{shapeshop}
F.~Hohman, N.~Hodas, and D.~H. Chau, ``Shapeshop: Towards understanding deep
  learning representations via interactive experimentation,'' in {\em
  Proceedings of the 2017 CHI Conference Extended Abstracts on Human Factors in
  Computing Systems}, pp.~1694--1699, ACM, 2017.

\bibitem{tensorPlayground}
D.~Smilkov, S.~Carter, D.~Sculley, F.~B. Vi{\'e}gas, and M.~Wattenberg,
  ``Direct-manipulation visualization of deep networks,'' {\em arXiv preprint
  arXiv:1708.03788}, 2017.

\bibitem{imageClassification}
{\'A}.~Cabrera, F.~Hohman, J.~Lin, and D.~H. Chau, ``Interactive classification
  for deep learning interpretation,'' {\em arXiv preprint arXiv:1806.05660},
  2018.

\bibitem{karpathy2015visualizing}
A.~Karpathy, J.~Johnson, and L.~Fei-Fei, ``Visualizing and understanding
  recurrent networks,'' {\em arXiv preprint arXiv:1506.02078}, 2015.

\bibitem{lstmvis}
H.~Strobelt, S.~Gehrmann, H.~Pfister, and A.~M. Rush, ``Lstmvis: A tool for
  visual analysis of hidden state dynamics in recurrent neural networks,'' {\em
  IEEE transactions on visualization and computer graphics}, vol.~24, no.~1,
  pp.~667--676, 2018.

\bibitem{cnn_vis}
M.~Liu, J.~Shi, Z.~Li, C.~Li, J.~Zhu, and S.~Liu, ``Towards better analysis of
  deep convolutional neural networks,'' {\em IEEE transactions on visualization
  and computer graphics}, vol.~23, no.~1, pp.~91--100, 2017.

\bibitem{seq2seq}
H.~Strobelt, S.~Gehrmann, M.~Behrisch, A.~Perer, H.~Pfister, and A.~M. Rush,
  ``Seq2seq-vis: A visual debugging tool for sequence-to-sequence models,''
  {\em arXiv preprint arXiv:1804.09299}, 2018.

\end{thebibliography}

\end{document}